\documentclass{article}
\usepackage{spconf,amsmath,graphicx, amsfonts}
\usepackage{enumitem}
\usepackage{booktabs, multirow}
\usepackage[pagebackref=true, colorlinks, urlcolor=black]{hyperref}

\begin{document}
\twocolumn[{%
\vspace{30mm}
{ \large
\begin{itemize}[leftmargin=2.5cm, align=parleft, labelsep=2cm, itemsep=4ex,]

\item[\textbf{Citation}]{G. Kwon, M. Prabhushankar, D. Temel, and G. AlRegib, “Novelty Detection Through Model-Based Characterization of Neural Networks,” 2020 27th \textit{IEEE International Conference on Image Processing (ICIP)}, Abu Dhabi, United Arab Emirates (UAE), 2020.}

\item[\textbf{Review}]{Date of Publication: October 25, 2020}

\item[\textbf{Codes}]{\url{https://github.com/olivesgatech/gradcon-anomaly}}

\item[\textbf{Bib}]  {@inproceedings\{kwon2020novelty,\\
    title=\{Novelty Detection Through Model-Based Characterization of Neural Networks\},\\
    author=\{Kwon, Gukyeong and Prabhushankar, Mohit and Temel, Dogancan and AlRegib, Ghassan\},\\
    booktitle=\{2020 IEEE International Conference on Image Processing (ICIP)\},\\
    year=\{2020\}\}}

\item[\textbf{Copyright}]{\textcopyright 2020 IEEE. Personal use of this material is permitted. Permission from IEEE must be obtained for all other uses, in any current or future media, including reprinting/republishing this material for advertising or promotional purposes,
creating new collective works, for resale or redistribution to servers or lists, or reuse of any copyrighted component
of this work in other works. }

\item[\textbf{Contact}]{
\{gukyeong.kwon, mohit.p, cantemel, alregib\}@gatech.edu\\
\url{https://ghassanalregib.info/}\\}
\end{itemize}
}}]
\newpage
\clearpage
\setcounter{page}{1}

% Title.
% ------
\title{Novelty Detection through Model-Based\\ Characterization of Neural Networks}
%
% Single address.
% ---------------
% \name{Author(s) Name(s)\thanks{Thanks to XYZ agency for funding.}}
% \address{Author Affiliation(s)}
\name{Gukyeong Kwon, Mohit Prabhushankar, Dogancan Temel, and Ghassan AlRegib}
\address{OLIVES at the Center for Signal and Information Processing,\\ School of Electrical and Computer Engineering,\\ Georgia Institute of Technology, Atlanta, GA, 30332-0250\\ \{gukyeong.kwon, mohit.p, cantemel, alregib\}@gatech.edu}
%
% For example:
% ------------
%\address{School\\
%	Department\\
%	Address}
%
% Two addresses (uncomment and modify for two-address case).
% ----------------------------------------------------------
%\twoauthors
%  {A. Author-one, B. Author-two\sthanks{Thanks to XYZ agency for funding.}}
%	{School A-B\\
%	Department A-B\\
%	Address A-B}
%  {C. Author-three, D. Author-four\sthanks{The fourth author performed the work
%	while at ...}}
%	{School C-D\\
%	Department C-D\\
%	Address C-D}
%
% \begin{document}
% \ninept
%
\maketitle
\begin{abstract}
In this paper, we propose a model-based characterization of neural networks to detect novel input types and conditions. Novelty detection is crucial to identify abnormal inputs that can significantly degrade the performance of machine learning algorithms. Majority of existing studies have focused on activation-based representations to detect abnormal inputs, which limits the characterization of abnormality from a data perspective. However, a model perspective can also be informative in terms of the novelties and abnormalities. To articulate the significance of the model perspective in novelty detection, we utilize backpropagated gradients. We conduct a comprehensive analysis to compare the representation capability of gradients with that of activation and show that the gradients outperform the activation in novel class and condition detection. We validate our approach using four image recognition datasets including MNIST, Fashion-MNIST, CIFAR-10, and CURE-TSR. We achieve a significant improvement on all four datasets with an average AUROC of $0.953$, $0.918$, $0.582$, and $0.746$, respectively.
\end{abstract}
\section{Introduction}
\label{sec:intro}
Characterization of novel data for machine learning algorithms has become an increasingly important topic for diverse applications including but not limited to visual recognition~\cite{hendrycks2019robustness}, speech processing~\cite{mitra2017robust}, and medical diagnosis~\cite{schlegl2017unsupervised}. In particular, when trained models are deployed in diverse environments \cite{Temel2018_CUREOR, Temel2019_CUREOR}, new classes of input (e.g. unknown objects) or conditions (e.g. inclement conditions such as rain and snow) \cite{Temel2018_SPM,temel2019traffic} that the models have not been exposed to during training can cause a significant performance degradation. To ensure the safety of machine learning algorithms in real-world scenarios, it is essential to characterize and detect novel data. 

\begin{figure}[t]
 	\centering
 	\includegraphics[width=.99\columnwidth]{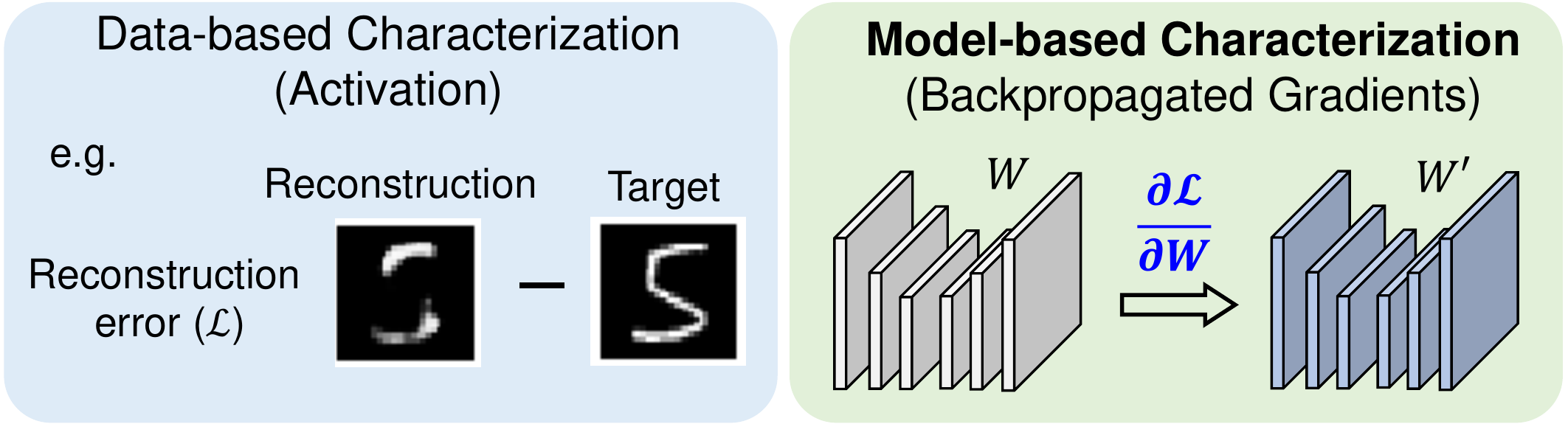}
 	\caption{Data-based and model-based characterization for novelty detection.}\label{fig:intro}\vspace{-0.5cm}
\end{figure}

Novelty detection, often also referred to as one-class classification or anomaly detection, is a research topic which aims to classify input data that is different in some aspects from training data~\cite{pimentel2014review}. A key element for the success of novelty detection is to learn a representation that can clearly separate normal and abnormal data. Most of existing works have focused on learning representations obtained in a form of activation. Novelty detection based on activation-based representations characterizes \textit{how much of the input corresponds to the learned information of the model}. For instance, assume that we input digit `5' (abnormal data) to an autoencoder trained to accurately reconstruct digit `0' (normal data). Based on the reconstructed image, which is the activation-based representation of the autoencoder, we calculate the reconstruction error as shown in the left side of Fig.~\ref{fig:intro}. Since the autoencoder has learned round shape information from `0', the curved edges at the top and the bottom of `5' are reconstructed but straight edges in the middle cannot be accurately reconstructed. The reconstruction error captures what the autoencoder has not learned and quantifies the abnormality. We can interpret this novelty detection based on activation-based representation as the characterization of abnormality from a data perspective.

In this paper, we propose to characterize novelty from a model perspective. In particular, we use backpropagated gradients from neural networks to obtain the model-based characterization of abnormality. A gradient is generated through backpropagation to train neural networks by minimizing designed loss functions~\cite{rumelhart1986learning}. The gradient with respect to the weights provides directional information to update the neural network. Also, the abnormal data requires more drastic updates on neural networks compared to the normal data. Therefore, by analyzing \textit{how much update of the model is required by data}, we can measure the novelty of input data as shown in the right side of Fig.~\ref{fig:intro}. We validate the effectiveness of gradient-based representations for novelty detection through comprehensive experiments. In particular, we compare the gradients with the activation-based representations and highlight the effectiveness of gradient features in novelty detection. Also, we perform novelty detection for different classes and conditions of inputs to show the generalizability of gradient features for different types of novelty. The contributions of this paper are three folds:
\begin{enumerate}[label=\roman*, leftmargin=0.5cm]
    \item We propose a framework to characterize novelty from the model perspective using gradients.
    \item We analyze the representation capability of the gradient compared to activations through controlled experiments.
    \item We validate the generalizability of gradient features for different classes and input conditions.
\end{enumerate}

\section{Related Works}
\label{sec:litearuatre}
\vspace{-0.2cm}Most of existing novelty detection algorithms are developed based on activation-based representations. During training, constraints are imposed to constrain the activation-based representation. For a query image, deviation of the representation from the constraints is measured as a novelty score for detection. In~\cite{sakurada2014anomaly, zhou2017anomaly}, the authors propose novelty detection algorithms using autoencoders based on the idea that the abnormal data is incompressible and cannot be accurately projected in the latent space. Therefore, they are poorly reconstructed and the errors can capture the anomaly in data. The authors in~\cite{zong2018deep} fit Gaussian mixture models (GMM) to reconstruction error features and latent variables and estimate the likelihood of inputs to detect novel data. In~\cite{sabokrou2018adversarially}, the authors adversarilally train a discriminator with an autoencoder to classify reconstructed images from original images and distorted images. The discriminator is utilized as a novelty detector during testing.

The usage of backpropagated gradients has not been limited to training neural networks. The backpropagated gradients have been widely used for visualization of deep networks. In~\cite{zeiler2014visualizing, springenberg2014striving}, information that networks have learned for a specific target class is mapped back to pixel space through the backpropagation and visualized. The authors in~\cite{selvaraju2017grad} utilize the gradient with respect to the activation to weight the activation and visualize reasoning for prediction that deep networks have made. An adversarial attack is another application of gradients. In~\cite{goodfellow2014explaining, kurakin2016adversarial}, the authors show that adversarial attacks can be generated by adding an imperceptibly small vector which is the signum of input gradients. In~\cite{kwon2019distorted}, the gradients of the neural network are utilized to classify distorted images and objectively estimate the quality of them. Several works have incorporated gradients with respect to input in a form of regularization during the training of neural networks to improve robustness~\cite{drucker1991double, ross2018improving}. Although existing works have shown that gradients with respect to the input or the activation can be useful for diverse applications, the gradients with respect to the weights of neural networks have not been actively explored as features for data characterization.\vspace{-0.2cm}

\section{Model-based Characterization}
\label{sec:gradients}
\begin{figure}[t]
\centering
\begin{minipage}[t]{0.32\linewidth}
  \centering
\includegraphics[width=\linewidth]{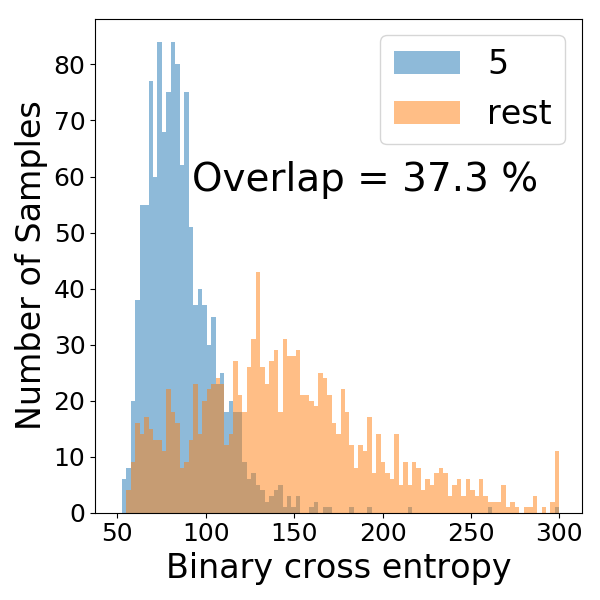}\vspace{-0.1cm}
  \centerline{\small{(a) Reconstruction error}}%\medskip
\end{minipage}
\begin{minipage}[t]{0.32\linewidth}
  \centering
\includegraphics[width=\linewidth]{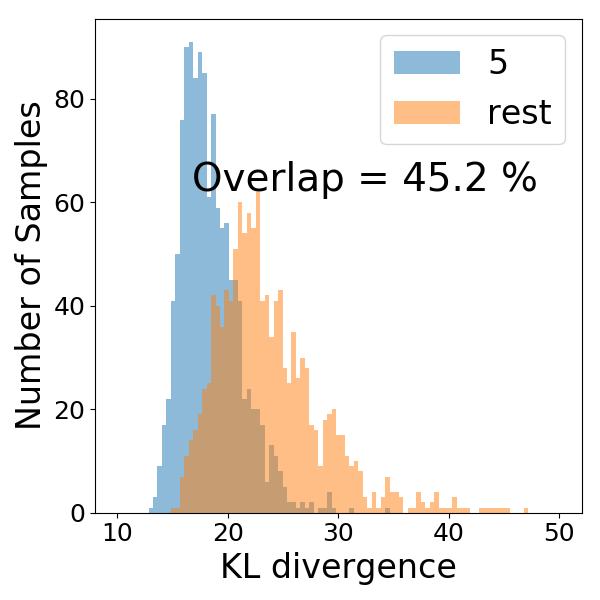}\vspace{-0.1cm}
  \centerline{\small{(b) Latent loss}}%\medskip
\end{minipage}
\begin{minipage}[t]{0.32\linewidth}
  \centering
\includegraphics[width=\linewidth]{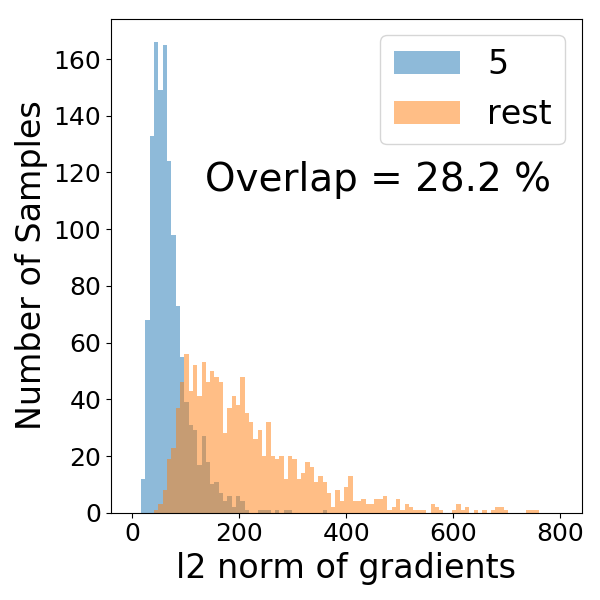}\vspace{-0.1cm}
  \centerline{\small{(c) Gradient}}%\medskip
\end{minipage}\vspace{-0.1cm}
\caption{Statistical deviation between inliers and outliers.}\label{fig:distributions}\vspace{-0.3cm}
\end{figure}

\vspace{-0.2cm}We use an autoencoder which is an unsupervised representation learning framework to explain the proposed approach of model-based novelty characterization. The autoencoder is trained to reconstruct inputs as outputs~\cite{goodfellow2016deep}. It consists of an encoder, $f_\theta$, and a decoder, $g_\phi$. From an input image, $x$, a latent variable, $z$, is generated as $z = f_\theta (x)$ and a reconstructed image is obtained by feeding the latent variable into the decoder, $g_\phi(f_\theta(x))$. The training is performed by minimizing a loss function, $J(x; \theta, \phi)$, defined as follows:\vspace{-0.1cm}
\begin{equation}\label{eq:loss}
    J(x; \theta, \phi) = \mathcal{L}(x, g_{\phi}(f_{\theta}(x))) + \Omega(z; \theta, \phi),\vspace{-0.1cm}
\end{equation}
where $\mathcal{L}$ is a reconstruction error which measures the dissimilarity between the input and the reconstructed image and $\Omega$ is a regularization term for the latent variable. In particular, when the latent space of the autoencoder is constrained to be Gaussian distribution, it becomes a variational autoencoder (VAE)~\cite{kingma2013auto}. The VAE has been an actively explored learning framework for novelty detection because it generates two well-constrained  activation-based representations, the latent representation and the reconstructed image representation~\cite{fan2018video}. Based on these constrained representations, the latent loss or the reconstruction error have been widely used as novelty scores by measuring the deviation of representations from the constraint. In our proposed framework, we extract backpropagated gradients from the trained VAE and use them as features to detect novelty. Given that the gradients guide updates required to the neural network by providing directional information, the gradients can capture the abnormality of query data from the perspective of model.

We perform statistical analysis on both activation-based and gradients to show the effectiveness of them in characterizing novel data. We train a VAE~\cite{kingma2013auto} by minimizing a loss defined as follows:\vspace{-0.1cm}
\begin{equation}
    J(x;\theta, \phi) = - \mathbb{E}_{g_\phi(z|x)}[\log f_\theta(x | z)] + \text{KL}[g_\phi(z|x)|| f(z)],
    \vspace{-0.1cm}
\end{equation}
where KL is the Kullback Leibler divergence between two distributions and we assume $f(z) = N(z| \textbf{0}, I)$. Therefore, KL divergence constrains the latent space of VAE to be the Gaussian distribution. The first term in the loss corresponds to the reconstruction error, $\mathcal{L}$, and calculated using binary cross entropy. The second term is the latent loss, $\Omega$ in~(\ref{eq:loss}). We train the VAE using digit `5' images in MNIST~\cite{lecun1998gradient} which are considered as inliers and other digit images are considered as outliers. We obtain the reconstruction error and the latent loss by passing test images through the VAE pre-trained with digit `5'. The gradients are extracted from the first layer of the decoder through the backpropagation of the reconstruction error from each test image. 

We visualize histograms of the reconstruction error, the latent loss, and the $\ell_2$ norm of gradients in Fig.~\ref{fig:distributions} (a), (b), (c), respectively. Furthermore, we provide percentages of samples in the overlapped region of the histograms to quantify the separation between two distributions from the inliers and the outliers. Ideally, large separation between the inliers and the outliers is desired for effective novelty detection. As shown in these histograms, the $\ell_2$ norm of backpropagated gradients, which measures the magnitude of gradients, better separates the inliers and the outliers than the reconstruction error and the latent loss. This shows that the magnitude of the gradients is more informative in characterizing abnormal data compared to other activation-based measures. In the following section, we utilize both magnitude and direction information of gradients by using them as features for novelty detection and highlight the performance from gradient features.\vspace{-0.2cm}

\section{Experiments}
\label{sec:experiments}
\begin{figure}[t]
 	\centering
 	\includegraphics[width=.8\columnwidth]{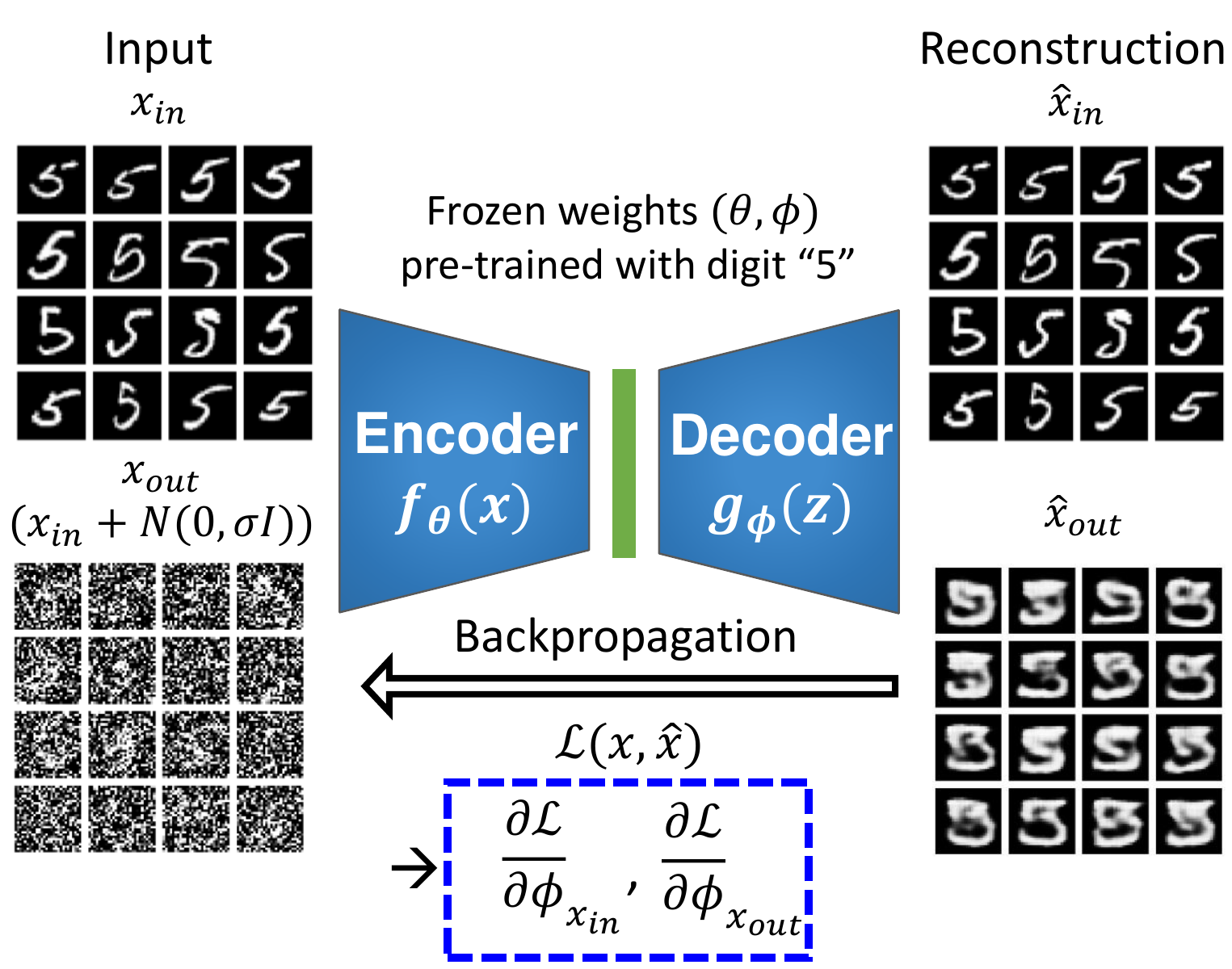}
 	\caption{Generation of gradient features.}\label{fig:framework}\vspace{-0.3cm}
\end{figure}
We conduct controlled experiments to further analyze the gradient features for novelty detection. In particular, we perform novel class detection and novel condition detection using gradients and compare the performance with activation-based representations. In novel class detection, samples from one class are considered as inliers and other class samples are considered as outliers. For novel condition detection, images without any effect are utilized as inliers and images captured under challenging conditions such as distortions or environmental effects are used as outliers. We only use the inliers for training and classify both inliers and outliers during testing. 

For the fair comparison between gradients and activation-based representations, we first train a VAE as described in Section~\ref{sec:gradients} using the inliers. Then, we train three different classifiers with the same architecture using reconstruction error, latent loss, backpropagated gradients as features. The classifier consists of two linear layers and sigmoid nonlinear activation layers between the linear layers. The reconstruction error and the latent loss are calculated as suggested in~\cite{kingma2013auto} but we do not take sum over all elements but use vectors as features for the classifiers. We obtain the gradients by backpropagating the reconstruction error as shown in Fig.~\ref{fig:framework}. To train a supervised classifier, we need outlier training images. As suggested in~\cite{sabokrou2018adversarially}, we distort the inlier images by adding Gaussian noise and use the distorted images as the training outliers. For the novel class detection, we extract gradients from the first layer of the decoder since the layer close to the latent representation is supposed to contain high-level semantic information. On the other hand, distortions or challenging conditions alter the low-level characteristics of images such as edges and colors. Therefore, we extract gradients from the last layer of the decoder for the novel condition detection. 

We use three image recognition datasets, which are MNIST~\cite{lecun1998gradient}, Fashion MNIST (fMNIST)~\cite{xiao2017fashion}, and CIFAR-10~\cite{krizhevsky2009learning}, for the novel class detection task and use CURE-TSR dataset~\cite{temel2017cure} for the novel condition detection. MNIST, fMNIST, and CIFAR-10 contain 10 classes of digits, fashion products and color objects, respectively. CURE-TSR contains traffic sign images with 12 challenging conditions and 5 challenge levels. We consider 5 challenging conditions which are \texttt{Lens blur}, \texttt{Dirty lens}, \texttt{Gaussian blur}, \texttt{Rain}, and \texttt{Haze} for this experiment. Test sets contain the same number of inliers and outliers. For MNIST and fMNSIT, we split the dataset into 5 folds and 60\% of each class is used for training, 20\% is used for validation, the remaining 20\% is used for testing. For CIFAR-10 and CURE-TSR, the original training and test splits are used and 10\% of the training images are held out for validation.\vspace{-0.2cm}

% \usepackage{multirow}
\begin{table*}[tb]
\small
\centering
\caption{Novelty class detection results on MNIST, fMNIST, and CIFAR-10.\label{tab:baseline}}
\begin{tabular}{ccccccccccccc}
% \hline\hline
\toprule
\multirow{2}{*}{Dataset} & \multirow{2}{*}{Repre.} & \multicolumn{10}{c}{Classes} & \multirow{2}{*}{Average} \\ \cline{3-12}
 &  & 0 & 1 & 2 & 3 & 4 & 5 & 6 & 7 & 8 & 9 &  \\ \hline\hline
\multirow{3}{*}{MNIST} & Recon. & 0.043 & 0.916 & 0.293 & 0.132 & 0.103 & 0.158 & 0.101 & 0.115 & 0.291 & 0.147 & 0.230 \\ \cline{2-13} 
 & Latent & 0.956 & 0.510 & 0.687 & 0.740 & 0.852 & 0.526 & 0.675 & 0.942 & 0.348 & 0.948 & 0.718 \\ \cline{2-13} 
 & Gradient & \textbf{0.985} & \textbf{0.994} & \textbf{0.941} & \textbf{0.928} & \textbf{0.953} & \textbf{0.926} & \textbf{0.980} & \textbf{0.960} & \textbf{0.894} & \textbf{0.968} & \textbf{0.953} \\ \hline\hline
\multirow{3}{*}{fMNIST} & Recon. & 0.778 & 0.952 & 0.831 & 0.799 & 0.801 & 0.787 & 0.748 & 0.939 & 0.610 & 0.932 & 0.818 \\ \cline{2-13} 
 & Latent & 0.733 & 0.642 & 0.525 & 0.877 & 0.715 & 0.831 & 0.585 & 0.961 & 0.702 & 0.835 & 0.741 \\ \cline{2-13} 
 & Gradient & \textbf{0.913} & \textbf{0.958} & \textbf{0.883} & \textbf{0.922} & \textbf{0.907} & \textbf{0.924} & \textbf{0.798} & \textbf{0.974} & \textbf{0.925} & \textbf{0.975} & \textbf{0.918} \\ \hline\hline
\multirow{3}{*}{CIFAR-10} & Recon. & 0.600 & 0.485 & 0.539 & \textbf{0.496} & 0.532 & 0.444 & 0.601 & \textbf{0.545} & 0.634 & 0.541 & 0.542 \\ \cline{2-13} 
 & Latent & \textbf{0.683} & 0.382 & 0.560 & 0.458 & 0.649 & 0.486 & \textbf{0.724} & 0.465 & \textbf{0.662} & \textbf{0.550} & 0.562 \\ \cline{2-13} 
 & Gradient & 0.658 & \textbf{0.543} & \textbf{0.632} & 0.461 & \textbf{0.725} & \textbf{0.493} & 0.699 & 0.490 & 0.641 & 0.477 & \textbf{0.582} \\ 
%  \hline\hline
\bottomrule
\end{tabular}
\end{table*}

\section{Results}
\label{sec:results}
\begin{figure*}[t]
\centering
\includegraphics[width=0.35\linewidth]{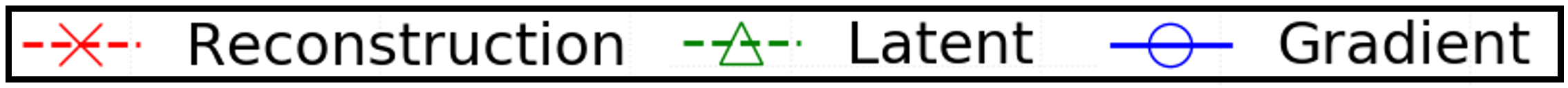}\\
\begin{minipage}[t]{0.195\linewidth}
  \centering
\includegraphics[width=\linewidth]{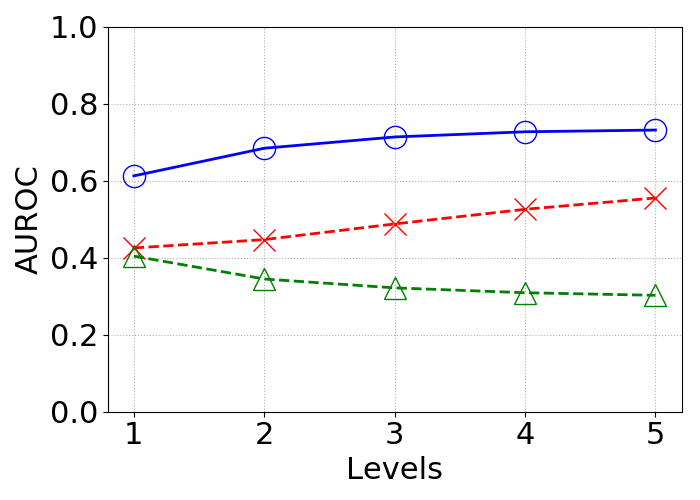}\vspace{-0.15cm}
  \centerline{\small{(a) Lens blur}}%\medskip
\end{minipage}
\begin{minipage}[t]{0.195\linewidth}
  \centering
\includegraphics[width=\linewidth]{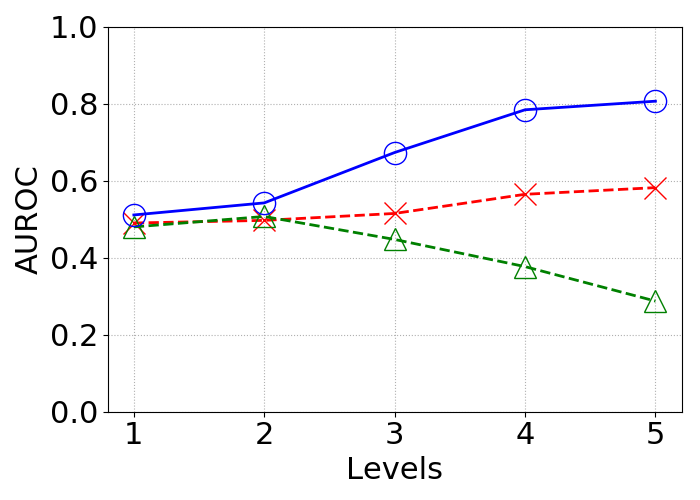}\vspace{-0.15cm}
  \centerline{\small{(b) Dirty lens}}%\medskip
\end{minipage}
\begin{minipage}[t]{0.195\linewidth}
  \centering
\includegraphics[width=\linewidth]{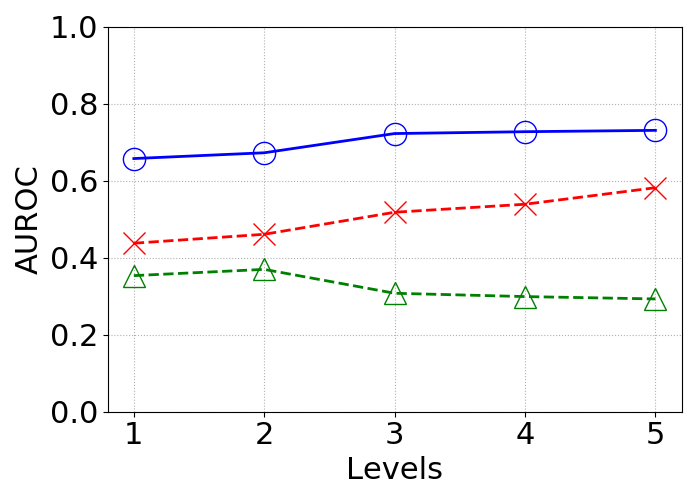}\vspace{-0.15cm}
  \centerline{\small{(c) Gaussian blur}}%\medskip
\end{minipage}
\begin{minipage}[t]{0.195\linewidth}
  \centering
\includegraphics[width=\linewidth]{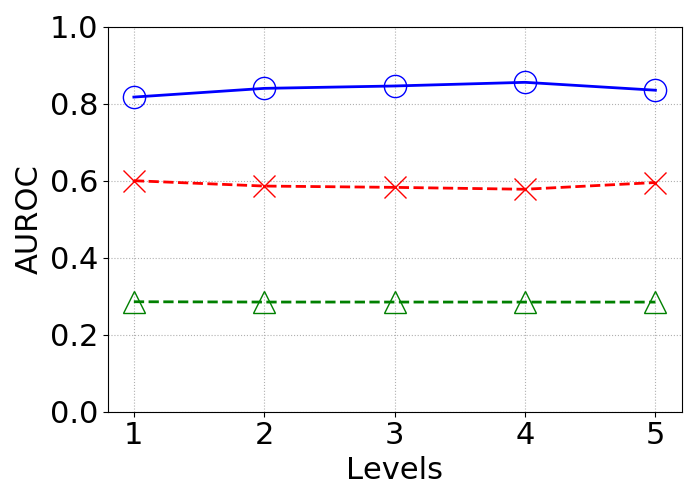}\vspace{-0.15cm}
  \centerline{\small{(d) Rain}}%\medskip
\end{minipage}
\begin{minipage}[t]{0.195\linewidth}
  \centering
\includegraphics[width=\linewidth]{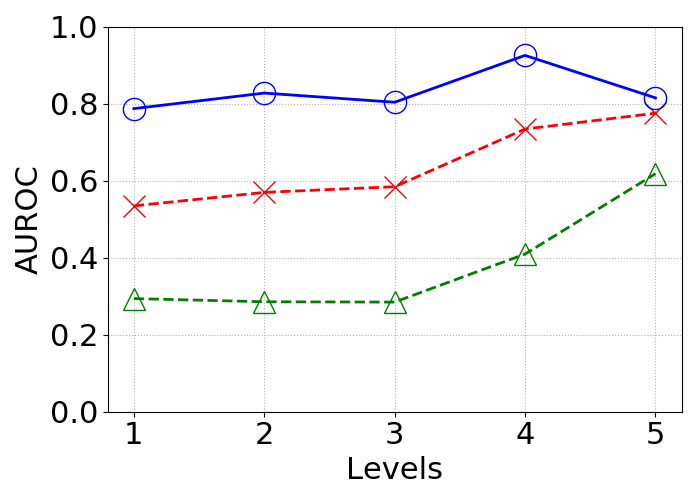}\vspace{-0.15cm}
  \centerline{\small{(e) Haze}}%\medskip
\end{minipage}
\caption{Novelty condition detection results on CURE-TSR.}\label{fig:cure_results}
\end{figure*}
\subsection{Novel class detection}
In Table.~\ref{tab:baseline}, we summarize the performance of the novelty class detectors trained using the activation-based representations (the reconstruction error and the latent loss) and the gradient. The performance is measured by area under receiver operation characteristic curve (AUROC) for each class and an average AUROC over different classes is also provided in the table. AUROC is bounded between 0 and 1 and the higher value indicates superior performance. As shown in the table, the classifiers trained on the gradients outperform those trained on the activation-based representations by a significant margin for almost all classes in three datasets. In particular, the best average AUROC performance obtained from the classifiers trained using the gradients is higher by $0.235$, $0.1$, and $0.02$ respectively compared to the second best results in three datasets. Also, the variances of AUROC over 10 classes obtained from the gradients are $0.001$, $0.003$ in MNIST and fMNIST, respectively. These variances are significantly smaller than the second smallest variances $0.044$ and $0.011$, respectively for both datasets. In CIFAR-10, the variance of AUROC by the gradients is the second smallest. This indicates that different classes of anomalies are separated and characterized robustly using the backpropagated gradients. The reconstruction error shows particularly low performance on MNIST and this may be resulted from the fact that digit images take relatively small portion of the entire image and the setup of using the distorted inliers does not help learning tight decision boundary around the inliers. Given that the reconstruction error is backpropagated to generate the gradient features, we can understand the significance of directional information from the gradients by comparing the performance of the reconstruction errors and the gradients. In all three datasets, the performance from the gradient features outperforms that from the reconstruction error by at least $0.04$ AUROC scores.\vspace{-0.3cm} 

\subsection{Novel condition detection}
In Fig.~\ref{fig:cure_results}, we visualize the AUROC results over different challenge levels in each challenge type using CURE-TSR. The classifiers trained using the gradients outperform those trained on the reconstruction error and the latent loss for all challenge types and challenge levels. In terms of an average AUROC over challenge levels, the gradient shows the largest improvement in \texttt{Rain} followed by \texttt{Lens blur} and \texttt{Gaussian blur}. When the challenge level is low, challenge images are similar to challenge-free images and hard to detect. Except for \texttt{Dirty lens}, the gradients achieve at least $0.187$ improvement over the second best results in all challenge types. Five challenging conditions are chosen to encompass acquisition imperfection, processing artifact, and environmental challenging conditions. The best results from the gradients show its representation capability in characterizing diverse types of challenging conditions.\vspace{-0.2cm}

\section{Conclusion}
\label{sec:conclusion}
In this paper, we proposed a framework to characterize abnormality from the model perspective using gradients. We conducted a comprehensive analysis to compare the performance of novelty detection from the activation and the gradient. The statistical analysis demonstrates that the larger separation between inliers and outliers is achieved using the gradients compared to the activation. Also, we shows that the classifiers trained using the gradients as features outperform those trained using common activation-based features in novel class and condition detection. Considering that most of existing works have only focused on developing descriptive activation-based representations, we leave using more sophisticated training schemes such as adversarial training with gradient features as remaining work for the future. 

\ninept
\bibliographystyle{IEEEbib.bst}
\bibliography{refs}

\end{document}